\documentclass[10pt,journal,compsoc]{IEEEtran}
\usepackage[noend]{algpseudocode}

\usepackage{algorithmicx,algorithm}

\usepackage{graphicx} % 需使用图片包
\usepackage{enumerate}
\usepackage{booktabs}
\usepackage{multirow}
\usepackage{pfnote}
\usepackage{amssymb}
\usepackage{bm}
\usepackage{lineno}
\usepackage{xcolor}
\usepackage{hyperref}
\hypersetup{
	colorlinks=true,
	linkcolor=black,
	filecolor=black,      
	urlcolor=black,
	citecolor=black,
}
\ifCLASSOPTIONcompsoc
  \usepackage[nocompress]{cite}
\else
  \usepackage{cite}
\fi

\ifCLASSINFOpdf
\else
\fi
\usepackage{amsmath}
\begin{document}
\bibliographystyle{IEEEtran}%规定参考文献的样式

\title{SpikeGraphormer: A High-Performance Graph Transformer with Spiking Graph Attention}

\author{Yundong Sun*, Dongjie Zhu, Yansong Wang, Zhaoshuo Tian, Ning Cao and Gregory O'Hare
        %John~Doe,~\IEEEmembership{Fellow,~OSA,}
        %and~Jane~Doe,~\IEEEmembership{Life~Fellow,~IEEE}% <-this % stops a space
\IEEEcompsocitemizethanks{

\IEEEcompsocthanksitem Y. Sun and Z. Tian are with the School of Astronautics, Harbin Institute of Technology, Harbin, China, 150001.
E-mail: \{hitffmy@163.com, tianzhaoshuo@126.com\} 
\IEEEcompsocthanksitem D. Zhu and Y. Wang are with the School of Computer Science and Technology, Harbin Institute of Technology, Weihai, China, 264209.
%\protect\\
E-mail: \{zhudongjie@hit.edu.cn, yansongwang0629@163.com\}

\IEEEcompsocthanksitem Ning Cao is with HebeiGejun Science \& Technology Company, Shijiazhuang, 050023, China.
E-mail: ning.cao2008@hotmail.com 
\IEEEcompsocthanksitem Gregory O'Hare is with the School of Computer Science \& Statistics, Trinity College Dublin, Dublin, Dublin 2, Ireland.
%\protect\\
E-mail: GREGORY.OHARE@tcd.ie

}% <-this % stops an unwanted space
\thanks{(Corresponding author: Yundong Sun, hitffmy@163.com)}}

\IEEEtitleabstractindextext{%
\begin{abstract}
Recently, Graph Transformers have emerged as a promising solution to alleviate the inherent limitations of Graph Neural Networks (GNNs) and enhance graph representation performance. Unfortunately, Graph Transformers are computationally expensive due to the quadratic complexity inherent in self-attention when applied over large-scale graphs, especially for node tasks. In contrast, spiking neural networks (SNNs), with event-driven and binary spikes properties, can perform energy-efficient computation. In this work, we propose a novel insight into integrating SNNs with Graph Transformers and design a Spiking Graph Attention (SGA) module. The matrix multiplication is replaced by sparse addition and mask operations. The linear complexity enables all-pair node interactions on large-scale graphs with limited GPU memory. To our knowledge, our work is the first attempt to introduce SNNs into Graph Transformers. Furthermore, we design SpikeGraphormer, a Dual-branch architecture, combining a sparse GNN branch with our SGA-driven Graph Transformer branch, which can simultaneously perform all-pair node interactions and capture local neighborhoods. SpikeGraphormer consistently outperforms existing state-of-the-art approaches across various datasets and makes substantial improvements in training time, inference time, and GPU memory cost (10 $\sim$ 20 $\times$ lower than vanilla self-attention). It also performs well in cross-domain applications (image and text classification). We release our code at \url{https://github.com/PHD-lanyu/SpikeGraphormer}.

\end{abstract}

}

\maketitle

\IEEEdisplaynontitleabstractindextext

% \fi
\IEEEpeerreviewmaketitle

\IEEEraisesectionheading{\section{Introduction}\label{sec:introduction}}


\IEEEPARstart{G}{raph}, consisting of nodes interconnected through edges, has become the knowledge base underpinning numerous intelligent applications, such as social networks \cite{jia2023srfa} and protein molecular networks \cite{wu2023cfago}. In recent years, Graph Neural Networks (GNNs) \cite{wu2019simplifying,kipf2016semi,velivckovic2018graph} that capitalize on message-passing mechanisms have demonstrated remarkable potential in graph representation learning. 
%These methods have significantly improved the performance of subsequent tasks and a wide array of intelligent applications by effectively capturing and encoding complex topological structures and relations within graphs. 
However, with more in-depth study of GNNs, some potential deficiencies of the message-passing mechanism are gradually revealed, such as heterophily \cite{zhu2020beyond}, over-squashing \cite{di2023over}, over-smoothing \cite{bose2023can}, bottlenecks in capturing long-range dependencies \cite{wu2022nodeformer}, and graph incompleteness\cite{ying2021transformers}, etc. Recently, a growing body of research has integrated Transformers into the field of graph representation learning, to overcome GNNs' inherent limitations \cite{wu2022nodeformer,ying2021transformers,wu2021representing,zhang2022hierarchical}. These novel approaches treat each node in the graph as an individual input token, and by employing a global attention mechanism, they can discern and capture subtle and potentially overlooked node relationships that may not be explicitly encoded within the graph's topology. This capability allows Graph Transformers to consider a broader context and more intricate dependencies between nodes, thereby enhancing their representation power for graph-based data analysis and learning tasks.
\begin{figure}[t]
  \centering
  \includegraphics[width=3.5in,trim=0 0 0 0]{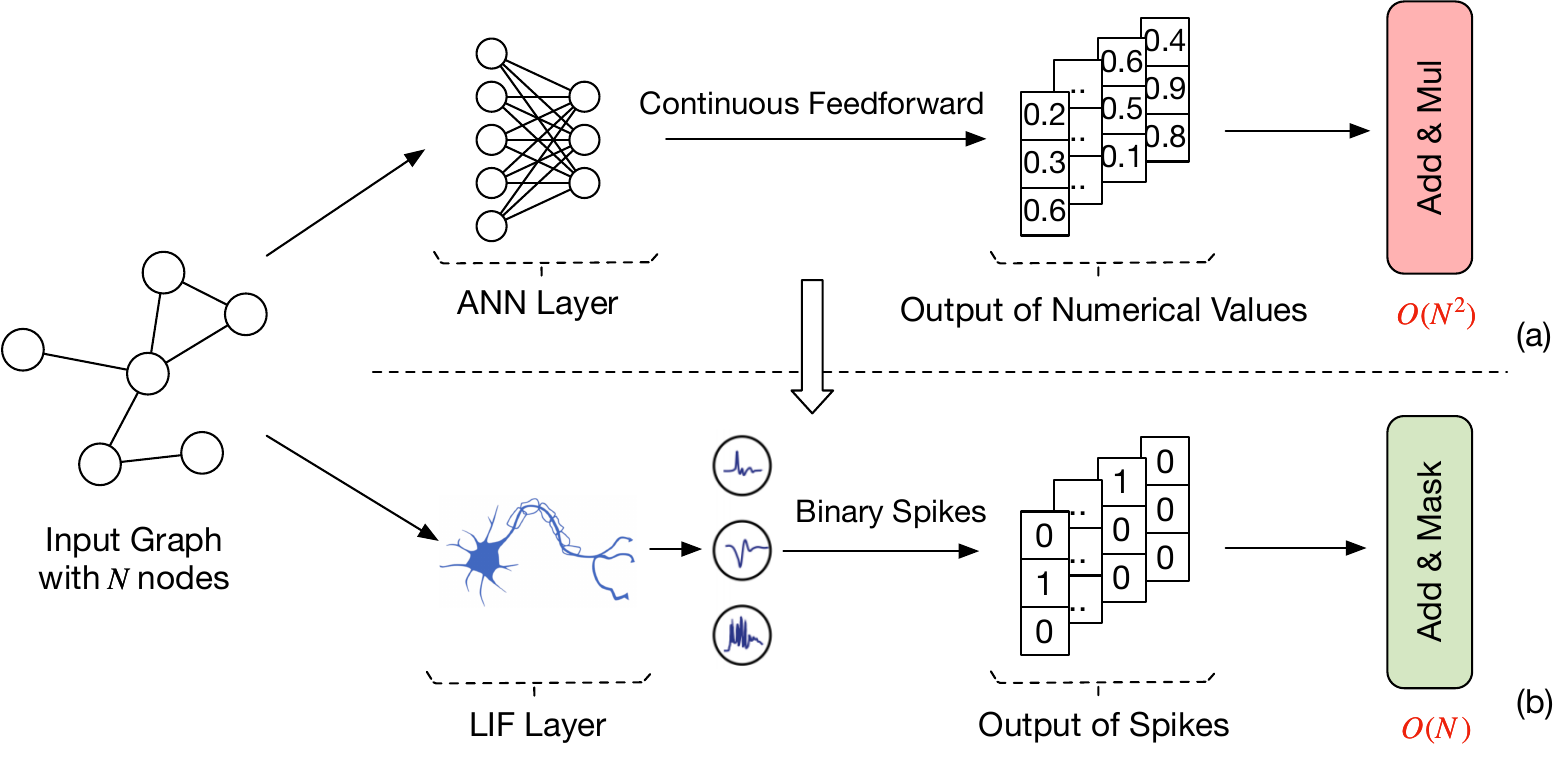}
  \caption{The diagram of (a) our Spike-based graph attention whose complexity is $O\left(N\right)$ vs (b) traditional ANN-based graph attention whose complexity is $O\left(N^2\right)$.}
  \label{fig1}
%  \vspace{-0.5cm}
  \end{figure}
  
  However, with the accumulation of data, the scale of the graph continues to expand, and the number of nodes is increasing. The quadratic computational complexity inherent in the global attention mechanism of Graph Transformers presents a headache performance bottleneck when dealing with node representation learning on large-scale graphs. This means that as the number of nodes in the graph increases, the computation and GPU memory consumption grow exponentially, which can limit the scalability and practical application of Graph Transformers for processing massive graph-structured data. Compared with graph-level prediction, node-level prediction needs to learn the embedding of each node. So, to achieve all-pair interactions, strategies such as graph coarsening \cite{zou2023dgsln} cannot be used to reduce the computational complexity. This brings a serious challenge to reduce the complexity of node-level prediction methods which want to maintain all-pair interaction capability.
  
  Several existing methods tackle this issue by focusing on selecting a subset of nodes for attention operations through various sampling techniques \cite{zhang2022hierarchical,zhao2021gophormer}. They aim to reduce the computational load while maintaining informative node representations. However, this scheme limits the model's attention view within these sampled nodes, which makes it fail to capture all-pair interactions and the global structure of the graph, deteriorating the expressive capacity of the Graph Transformer. Nodeformer \cite{wu2022nodeformer}, a recent study, introduces a novel approach involving kernelized messages passing to approximate all-pair attention in a more efficient manner with linear complexity. While this presents a promising avenue for scaling graph neural networks, it incorporates random feature maps which may bring instability during the training process.
  
  As third-generation neural networks, spiking neural networks (SNNs) \cite{yao2023attention}, with unique event-driven and binary spikes characteristics, can provide more energy-efficient computing and are likely to be a low-energy alternative to traditional artificial neural networks (ANNs) in the future. In light of these potential benefits, this paper explores how to harness SNNs to implement graph attention mechanisms. Leveraging SNNs' inherent low-energy characteristics, \textit{we investigate whether it is feasible to design a Graph Attention module based on SNNs that can effectively address the performance bottleneck associated with node representation learning on large-scale graphs.} By doing so, we aim to bridge the gap between energy efficiency and scalability while maintaining or even enhancing the quality of learned node representations in complex graph structures.
  
  To solve this problem, we propose the Spiking Graph Attention module, which employs spiking neuron layers to transform the numerical inputs $Q$, $K$, and $V$ into binary spikes. Subsequently, the complex matrix multiplication operation of attention computation can be replaced by the sparse addition and mask operations, so the quadratic complexity can be reduced to linear complexity. (from Fig.\ref{fig1}(a) to Fig.\ref{fig1}(b)). At the same time, we remove softmax and scale, because matrix column-wise summation and spiking neuron layer can play an equivalent role.
  
  At the same time, we design a Dual-branch architecture (SpikeGraphormer), which tackles the inherent limitations of the Transformer in capturing local graph structure by incorporating a sparse GNN branch that is adept at modeling graph structures and capturing the local neighborhoods of nodes. On this basis, we leverage a simple yet surprisingly effective fusion strategy (summation or concatenation), to integrate the output of the GNN and Transformer branches.
  
  The key contributions of this paper can be summarized as follows:

  %\subsection{Units}
\begin{itemize}
\item \textbf{Novel Insights:}

We propose a novel and valuable insight into the potential of integrating bio-inspired SNNs with Graph Transformers. The event-driven and binary-spike properties can greatly improve the computational efficiency of Graph Transformer, making it possible to realize all-pair node interactions on large-scale graphs with limited GPU memory. To our knowledge, our work is the first attempt to introduce SNNs into Graph Transformers.

\item \textbf{Linear Complexity Improvement:}

Based on our novel insights, our well-designed Spiking Graph Attention (SGA) module can fully leverage the bio-fidelity inference process of SNNs to replace the complex matrix multiplication operation of vanilla self-attention with sparse addition and mask operations. So, the overall computation can be optimized from quadratic to linear complexity of the number of nodes.

\item \textbf{Dual-branch Architecture Design: }

To eliminate the Transformers' deficiency in capturing local graph structures, we design SpikeGraphormer, a Dual-branch architecture, combining a sparse GNN branch with our proposed SGA-driven Graph Transformer branch. The simple yet surprisingly effective fusion strategy enables SpikeGraphormer to simultaneously perform all-pair node interactions and capture local neighborhoods, making it comparable to or even better than the latest ANN-driven Graph Transformer.

\item \textbf{Comprehensive Experiment Validation:}

Through extensive experimentation across various datasets with differing scales, we demonstrate that SpikeGraphormer achieves state-of-the-art performance while significantly reducing training and inference time, as well as GPU memory consumption (10 $\times$ $\sim$ 20 $\times$  lower than vanilla self-attention), especially on large-scale graphs. The excellent performance in image and text classification also reveals its great cross-domain generalization. We believe that our work can provide a valuable reference for subsequent research aiming at exploring SNNs for high-performance graph representation learning.

\end{itemize}

The remainder of this paper is organized as follows: Section \ref{sec2} will discuss some research works related to this paper. We will introduce the overall architecture and details of our proposed method, and give an in-depth analysis in Section \ref{sec3}. We compare SpikeGraphormer with existing methods in Section \ref{sec4}. In Section \ref{sec5}, we describe the experiment process and analyze the experimental results. Section \ref{sec6} concludes the paper and investigates some future research directions.

\section{Related Work}\label{sec2}

\subsection{Graph Neural Networks and Graph Transformers}\label{sec2.1}

Relying on convincing performance, GNNs have become a powerful weapon for graph data representation and analysis. The message-passing mechanism is the core of most GNNs, which fits the phenomenon that neighbor nodes in a graph are more likely to share similar attributes or labels due to their interconnection \cite{guo2023taming}. However, its expressive power is bounded by the Weisfeiler-Lehamn isomorphism hierarchy \cite{zhang2023complete}. Worse still, some inherent limitations of GNNs have been revealed, such as heterophily \cite{zhu2020beyond}, over-squashing \cite{di2023over}, over-smoothing \cite{bose2023can}, long-range dependencies \cite{wu2022nodeformer}, and graph incompleteness \cite{ying2021transformers}. 

To solve the above problems of GNNs, researchers have begun to leverage Transformers for graph representation \cite{wu2022nodeformer,ying2021transformers,wu2021representing,zhang2022hierarchical}. These methods treat each node as a token and feed it into the self-attention module for information exchange. At the same time, structural information is integrated into the Transformer by introducing node position embedding or injecting adjacency matrixes into attention calculation. The self-attention mechanism in Graph Transformers brings a global interaction capability that not only addresses the GNNs' limitations but also can capture implicit dependencies such as long-range interactions or unobserved potential links in the graph.

Unfortunately, the quadratic complexity of vanilla self-attention makes existing methods fail to scale to larger graph data with limited GPU memory. Therefore, the latest researches are devoted to mitigating the computational complexity of Graph Transformers with the hope of obtaining salable Graph Transformers.

\subsection{Salable Graph Transformers}\label{sec2.2}

One of the straightforward strategies is to obtain a subset of nodes through node sampling and feed them into the self-attention module, to alleviate computational load \cite{zhang2022hierarchical}. Some more extreme methods limit the attention scope to the local neighborhood of the central node (such as ego-graph) \cite{zhao2021gophormer}. Another kind of method conducts graph coarsening or node-dropping to reduce the number of nodes \cite{zhu2023hierarchical}. However, graph coarsening or node-dropping operations will bring about computational costs and information loss from the original graph. These methods are generally limited to graph representation and are not friendly to node representation tasks. There is also a novel work \cite{chen2022nagphormer} that aggregates the neighboring features of neighboring nodes within the same hop distance and treats them as individual tokens. By doing this, each token encapsulates information about a specific neighborhood radius around a target node, and the number of tokens is reduced to hop+1 (including the central node: 0-hop). However, this extremely coarse-grained processing method will lose a lot of information, and it will not be able to achieve all-pair interactions.

Some recent research has begun to focus on optimizing attention algorithms. Nodeformer \cite{chen2022nagphormer} is an innovative approach that proposes a kernelized message-passing scheme to approximate the all-pair attention mechanism with linear computational complexity. However, the introduction of random feature maps is a trade-off between computational efficiency and model stability. Therefore, in the face of the increasing data scale, how to develop a low-complexity Graph Transformer that can be extended to large-scale graphs is an urgent problem to be solved.

\subsection{Bio-inspired Spiking Neural Networks}\label{sec2.3}

In contrast to ANNs, which rely on iterative optimization of continuous numerical parameters, SNNs are designed to emulate the information processing and transmission mechanisms inherent in biological neural systems. At a given time, the SNN receives a series of binary spikes as input. If the accumulated or integrated signal at a neuron surpasses a predefined threshold, that neuron generates and propagates a binary spike event to its connected downstream neurons \cite{yao2023attention}. Event-driven and binary spikes allow SNNs to work rapidly. At the same time, more energy-efficient computation can be achieved by deploying SNNs on neuromorphic chips \cite{rathi2023exploring}. In recent years, some researchers have tried to integrate SNNs with deep learning technology, which greatly improves the efficiency of models on specific tasks. For example, spike-driven Transformer \cite{yao2024spike} and Spikformer \cite{zhou2022spikformer} incorporate the spik-driven paradigm into the Transformer and apply it to computer vision. Under the premise of greatly reducing computation energy, they can achieve satisfactory performance. Facing the automatic speech recognition (ASR) task, DyTr-SNN \cite{wang2023complex} introduces four types of neuronal dynamics to post-process the sequential patterns generated from the spiking transformer. The experimental results show that SNNs exhibit considerable potential for application in ASR tasks.

Although many studies have applied SNNs to various deep learning tasks and significantly reduced computation energy, no relevant studies have applied them to Graph Transformers. With the increasing scale of graph data, capturing all-pair node interactions on large-scale graphs faces an unavoidable efficiency bottleneck. In this paper, we investigate whether it is feasible to design a Graph Attention model based on SNNs that can effectively address the performance bottleneck. In this way, we aim to bridge the gap between energy efficiency and scalability while maintaining or even enhancing the quality of learned node representations in complex graph structures.

\begin{figure*}
  \centering
  \includegraphics[width=6in,trim=0 0 0 0]{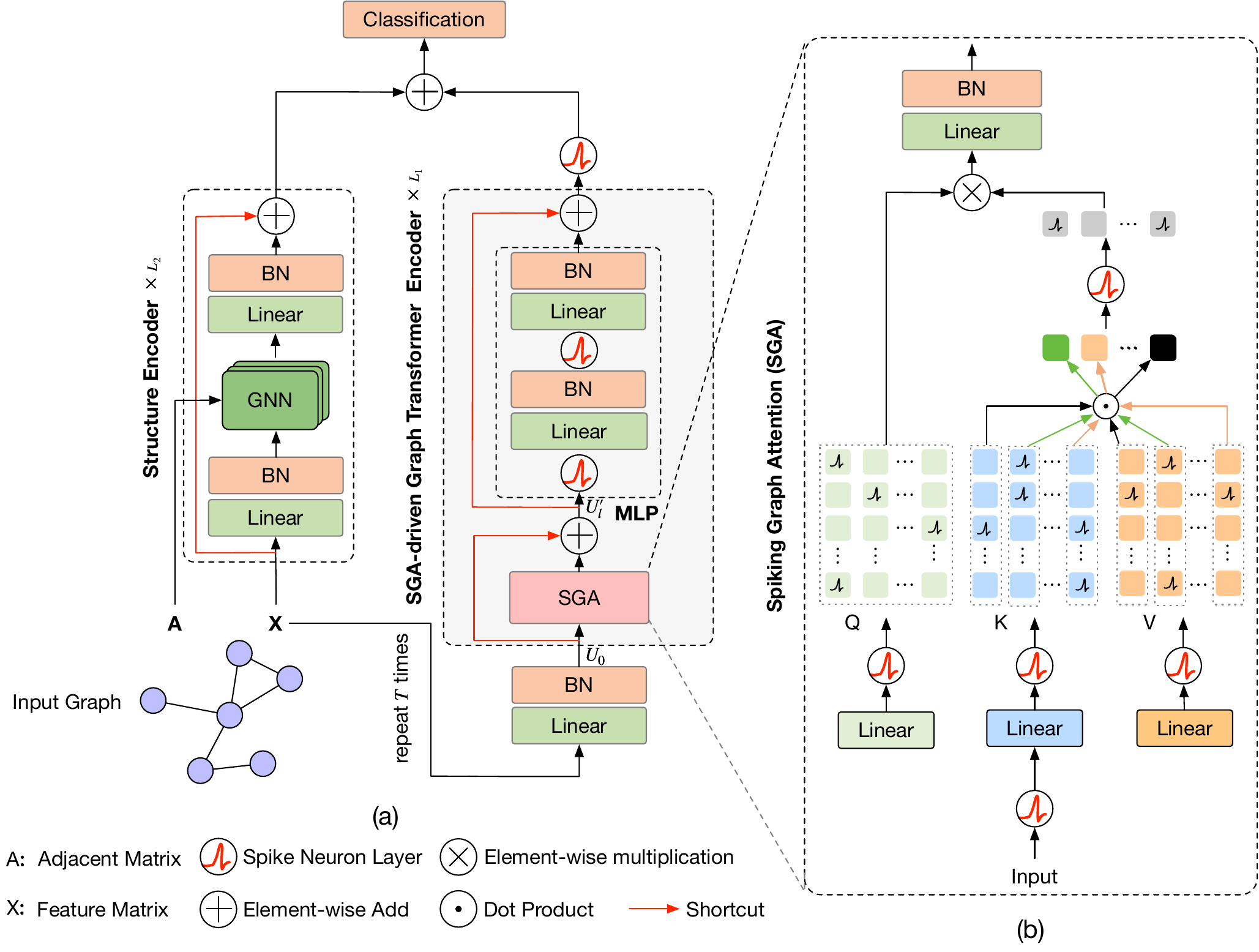}
 \caption{The overall architecture of SpikeGraphormer and the details of Spiking Graph Attention.}
  \label{fig2}
%  \vspace{-0.5cm}
  \end{figure*}

\section{SpikeGraphormer}\label{sec3}

In this paper, we propose a novel strategy for integrating SNNs and Graph Transformers. Specifically, we design SpikeGraphormer, a Dual-branch architecture, combining a sparse GNN branch with our proposed SGA-driven Graph Transformer branch. Before delving into the overall architecture and core components of SpikeGraphormer, we first introduce the spike neuron layer.

\subsection{Spiking Neuron Model}\label{sec3.1}

In this paper, we adopt the Leaky Integrate-and-Fire threshold spiking neuron (LIF) \cite{maass1997networks}, which is the most commonly used SNN neuron. The LIF neuron model strikes an optimal balance between the intricate spatio-temporal dynamics inherent in biological neurons and the necessitated mathematical simplifications, rendering it a fitting choice for simulating large-scale SNNs. It can be described by a differential function:

\begin{equation}
	\tau \frac{du(t)}{dt}= -u(t)+I(t)	
\end{equation}

In this context, $\tau$ represents the time constant, while $u\left(t\right)$ signifies the membrane potential of the postsynaptic neuron, and $I\left(t\right)$ denotes the aggregate input received from the ensemble of presynaptic neurons at any given time. After solving this differential equation, a simple iterative representation of the LIF-SNN layer\cite{yao2023attention,yao2024spike} is governed by:

\begin{equation}
\begin{split}
 	 &U[t]=H[t-1]+X[t],  	\\
 	 &S[t]=Hea(U[t]-u_{th}),  \\
 	 &H\left[t\right]=V_{reset}S\left[t\right]+\left(\beta U\left[t\right]\right)\odot\left(1-S\left[t\right]\right)
	\end{split}
\end{equation}
where $t$ denote the time step, $X\left[t\right]$ is the spatial input information and $H\left[t-1\right]$ is the temporal input. After coupling $X\left[t\right]$ and $H\left[t-1\right]$, we can obtain $U\left[t\right]$, which represents the membrane potential, $u_{th}$ is the threshold. The value of $X\left[t\right]$ can be derived through a variety of computational operators, including Convolutional (Conv) operations, Multilayer Perceptrons (MLP), and Self-Attention. $Hea\left(\cdot\right)$ is a Heaviside step function, if $x \geq 0$, $Hea\left(x\right)=1$, otherwise $Hea\left(x\right)=0$. Therefore, the spatial output tensor $S\left[t\right]$ could only be a 1 or a 0. $H\left[t\right]$ means the temporal output, $V_{reset}$ denotes the reset potential which is set after activating the output spiking. $\beta=e^{-\frac{dt}{\tau}}<1$ is the decay factor, $\odot$ is the element-wise multiplication.

\subsection{Overall Architecture}\label{sec3.2}

Fig.\ref{fig2}(a) shows the overall architecture of SpikeGraphormer, which includes two branches, sparse GNN and SGA-driven Graph Transformer branch. The SGA-driven Graph Transformer is the core branch of SpikeGraphormer. It takes the node feature matrix $X$ as the only input after $T$ repeats and leverages $L_1$ encoder blocks for all-pair node interactions.

The core module of the SGA-driven Graph Transformer branch is our elaborately designed Spiking Graph Attention (SGA), the detailed structure is shown in Fig.\ref{fig2} (b), and we will describe it in detail in Section \ref{sec3.3}. We also designed the GNN branch shown in the left part of Fig.\ref{fig2} (a), which can better capture the graph structure and collaborate with the SGA-driven Graph Transformer branch to obtain richer information. We will provide a detailed description of the GNN branch in Section \ref{sec3.4}.

The overall calculation process can be described by the following equations:

\begin{equation}
\begin{split}
 	 &S_0=\mathcal{SNN}\left(U_0\right),S_0\in R^{T\times N\times D},  	\\
 	 &U_l^\prime=SGA\left(S_{l-1}\right)+U_{l-1},U_l^\prime\in R^{T\times N\times D},l=1\ldots L_1,  \\
 	 &S_l^\prime=\mathcal{SNN}\left(U_l^\prime\right),S_l^\prime\in R^{T\times N\times D},l=1\ldots L_1, \\
 	 &S_l=\mathcal{SNN}\left(MLP\left(S_l^\prime\right)+U_l^\prime\right),S_l=R^{T\times N\times D},l=1,\ldots L_1
	\end{split}
\end{equation}
where $U_0\in R^{T\times N\times D}$, if the graph is static, that is, $X\in R^{N\times D}$ is a static matrix, then the node feature matrix $X$ is obtained by Linear and BN networks after $T$ repetitions. $N$ represents the node number of the input graph, $D$ is the feature dimension, $\mathcal{SNN}$ indicates the spike neuron layer, $T$ is the time steps of $\mathcal{SNN}$. To improve the ability of the model to capture different order features, we add residual structure to both SGA and MLP. After the features are passed through $L_1$ blocks of the SGA-driven Graph Transformer encoder, we can obtain the binary feature representation $S_{L_1}$.

\subsection{Spiking Graph Attention}\label{sec3.3}

Fig.\ref{fig2}(b) shows the detailed network structure of Spiking Graph Attention. First, we obtain $Q$, $K$, and $V$ through the SNN layer and linear network:

\begin{equation}
\begin{split}
 	 &U_0^\prime=\mathcal{SNN}\left(U_0\right),  	\\
 	 &Q=Liner_q\left(\mathcal{SNN}\left(U_0^\prime\right)\right),  \\
 	 &K=Liner_k\left(\mathcal{SNN}\left(U_0^\prime\right)\right), \\
 	 &V=Liner_v\left(\mathcal{SNN}\left(U_0^\prime\right)\right)
	\end{split}
\end{equation}

$U_0^\prime$ obtained after $U_0$ passes through $\mathcal{SNN}$, so $U_0^\prime$ a binarized spike tensor, $Liner_q$, $Liner_k$, and $Liner_v$ can be implemented by addition operation, which has extremely high computational efficiency.

In practical applications, the calculation of Graph Attention can adopt multiple channels. Following \cite{yao2024spike}, we set the number of channels $H=D$, and the results of different channels can be concatenated. The whole calculation process can be expressed by the following equation:

\begin{equation}
\scriptsize
SGA\left(Q^i,K^i,V^i\right)=\mathcal{SNN}\left(Q^i\right)\left(\mathcal{SNN}\left(K^i\right)^\top\odot\mathcal{SNN}\left(V^i\right)\right)
\end{equation}
where $Q^i$, $K^i$, and $V^i$ are $i$-th columns of the matrix $Q$, $K$ and $V$, respectively. In particular, it is important to note that $\mathcal{SNN}\left(Q^i\right)^\top\odot\mathcal{SNN}\left(V^i\right)$ yields a scalar, which is either 0 or 1. Therefore, the calculation of the whole equation (5) can be regarded as a mask operation, the dimension of vector $Q^i$, $K^i$ and $V^i$ is $N$, the complexity is $O\left(N\right)$. For all columns of the matrix $Q$, $K$, and $V$, we only need to concatenate each column after equation (5). Therefore, the complexity of the entire attention calculation is $O\left(ND\right)$, which is the linear complexity of the number of nodes $N$ and the node feature dimension $D$. What's more, it can be seen from the analysis of \cite{yao2024spike} that $K^i$ and $V^i$ are very sparse, and the sparsity is generally less than $0.01$. Therefore, the number of addition calculations for the entire SGA is generally less than $0.02 ND$, which is a very low-energy operation.

\begin{figure}[t]
  \centering
  \includegraphics[width=3.5in,trim=0 0 0 0]{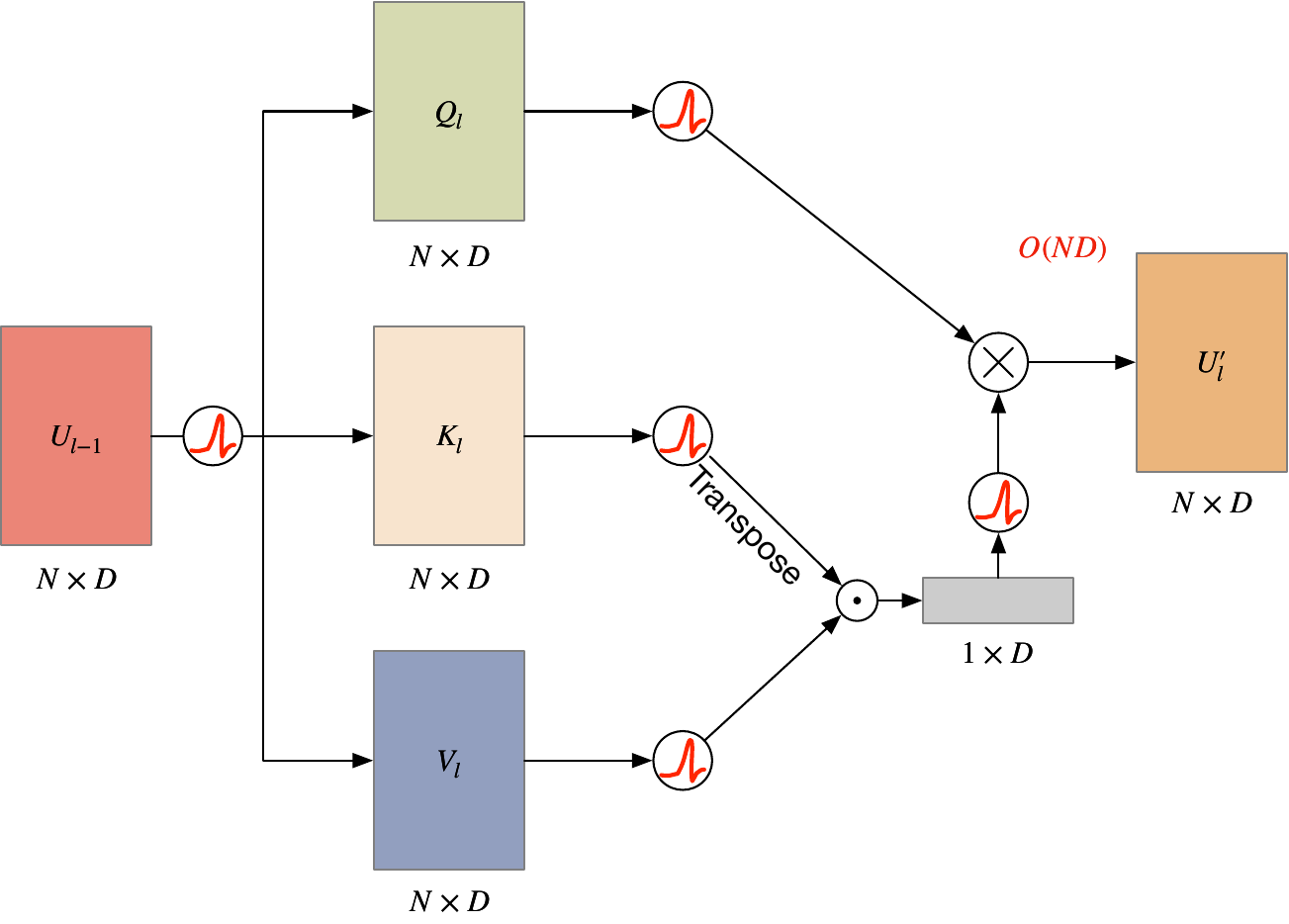}
  \caption{The feed-forward process of SGA from matrix views.}
  \label{fig3}
%  \vspace{-0.5cm}
  \end{figure}

In the practical implementation,  all columns of $Q$, $K$, and $V$ of all nodes can be performed simultaneously through matrix parallel operations. More importantly, highly concurrent matrix operations can better exert the computing power of GPU, which is particularly beneficial when dealing with large graphs. To more intuitively show the implementation details of SGA, we show the feed-forward process of SGA from matrix views in Fig.\ref{fig3}.

\begin{table*}[] 
\center
  \caption{The comparison between SpikeGraphormer and existing Graph Transformers w.r.t. target tasks (graph-level or node-level prediction), required architecture (PE for positional embeddings, AL for additional training loss, EE for edge embeddings), all-pair interaction capability and complexity w.r.t. node number $N$ and edge number $E$ (often $E\approx N\ll N^2)$.}
\label{tab1}
\begin{tabular}{c|c|ccc|c|cc}
\toprule

\multirow{2}{*}{\textbf{Model}}                    & \multirow{2}{*}{\textbf{Task}}   & \multicolumn{3}{c|}{\textbf{Architecture}} & \textbf{All-pair Node}  & \multicolumn{2}{c}{\textbf{Complexity}}                        \\
                         &      & \textbf{PE}        & \textbf{AL}        & \textbf{EE}       &   \textbf{Interactions}                         & \multicolumn{1}{c}{\textbf{Pre-processing}}         & \multicolumn{1}{c}{\textbf{Training}} \\                          
           \midrule
                         
GraphTransformer \cite{dwivedi2020generalization} & GL   	& R  	& - 		& R     & Yes   & $O\left(N^3\right)$ & $O\left(N^2\right)$
                           \\

Graphormer \cite{ying2021transformers}      & GL   & R     & -     & R     & Yes   & $O\left(N^3\right)$   &  $O\left(N^2\right)$                          \\

GraphTrans \cite{wu2021representing}       & GL   & -         & -         & -        & Yes                        & -                      & $O\left(N^2\right)$                     

      \\

GraphGPS \cite{rampavsek2022recipe}        & GL   & R         & -         & R        & Yes                        & $O\left(N^3\right)$ & $O\left(N+E\right)$ 
                          \\
                        
                          \midrule

ANS-GT{[} \cite{zhang2022hierarchical}           & NL   & R         & -         & -        & No                         & -                      & $O\left(Nsm^2\right)$                           \\
Nodeformer \cite{wu2022nodeformer}        & NL   & R         & R         & -        & Yes                        & -                      & $O\left(N+E\right)$                           \\
\textbf{SpikeGraphormer}(ours)    & NL   & -         & -         & -        & Yes                        & -                      & $O\left(N+E\right)$  
\\ 
\midrule
                    
\end{tabular}

\end{table*}

\subsection{Dual-branch architecture: Incorporation of Graph Structural Information}\label{sec3.4}

Although the SGA-driven Graph Transformer branch proposed in this paper can model all-pair node interactions, it does not contain the graph structure information, which may cause information loss. To incorporate prior structure information inherent in the input graph, prevailing models often resort to positional embedding \cite{chen2022structure}, edge regularization losses \cite{wu2022nodeformer}, injecting GNNs into Transformer architecture\cite{wu2021representing}, or optimizing attention by graph adjacent matrix \cite{min2022neighbour}. In this paper, we leverage a simple yet surprisingly effective fusion strategy (addition or concatenation), to integrate the output of the designed sparse GNN and SGA-driven Graph Transformer branch. Here, we take addition as an example:

\begin{equation}
	Z=\left(1-\alpha\right)S_l+\alpha GNN\left(X,A\right)
\end{equation}
where $\alpha$ represents the weight of the sparse GNN branch (abbreviated as graph weight), which is a manually determined hyperparameter. We can select different GNN structures according to the actual scenario, such as GAT \cite{velivckovic2018graph}, GCN \cite{kipf2016semi}, etc. In this paper, we choose the simplest GCN \cite{kipf2016semi}. It is worth mentioning that in the implementation of this paper, sparse matrix multiplication is leveraged, whose complexity is controlled in linear space $O(E)$. Fortunately, in real-world scenarios, most graphs are "sparse": the number of edges tends to be linear to the number of nodes. Therefore, it also remains within the linear complexity of the number of nodes.

Finally, the obtained node representation $Z$ is input into the predictor for the downstream task. In this paper, we select the node classification as the objective task, and the simplest linear network is selected as the classifier:

\begin{equation}
Y=CLS\left(Z\right)	
\end{equation}

\subsection{Complexity Analysis}\label{sec3.5}

As analyzed above, the overall complexity of our method includes two parts: (1) The complexity of the SGA-driven Graph Transformer branch is $O\left(ND\right)$; (2) The complexity of the GNN branch is $O\left(E\right)$. So the overall complexity is $O\left(ND+E\right)$. Due to the sparsity of the graph, $E\approx N\ll N^2$, $D\ll N$, therefore, the final complexity can be equivalent to $O(N+E)$, which is also the linear complexity of the number of nodes. In conclusion, the low-complexity, lightweight structure of the method proposed in this paper allows it to possess a high computational efficiency.

\subsection{Scaling to Larger Graphs}\label{sec3.6}

In the field of graph representation learning, the exponential increase in GPU memory cost w.r.t. the number of graph nodes is an ongoing problem. So, even the most light models, like GCN \cite{kipf2016semi}, cannot be trained in a full-batch manner on a single GPU when facing large-scale graphs. To solve this problem, we leverage graph sampling and mini-batch training technology \cite{wu2024simplifying}. Similarly, under the mini-batch training paradigm, our method maintains linear complexity and a larger batch size can be set. As a result, a broader range of all-pair node interactions can be effectively modeled within each batch, thereby capturing more comprehensive global features. There is no doubt that our method is orthogonal to other training optimization techniques, such as adaptive inference \cite{pope2023efficiently,zheng2023adaptive} and node neighbor sample \cite{zhao2023learnable,yang2020understanding}. In the future, we can combine SpikeGraphormer with these methods.

\section{Comparison with Existing Models}\label{sec4}

Table \ref{tab1} presents a comparative analysis of SpikeGraphormer against several state-of-the-art Graph Transformers in terms of target tasks, required architecture, all-pair interaction capability, and complexity. From the perspective of model architecture, most of the existing methods need to integrate the graph structure information by position embedding and edge embeddings, or introduce additional training tasks, such as edge regularization \cite{wu2022nodeformer}, which increase model complexity and overhead. We design a simple but effective dual-branch architecture and information fusion strategy to ensure the lightweight structure of SpikeGraphormer. While some graph-level prediction methods, such as GraphGPS \cite{rampavsek2022recipe}, exhibit a relatively low computational complexity during the training phase, their extremely complex preprocessing methods also result in high overall complexity. For the node-level prediction methods, it is very challenging to reduce the complexity while maintaining all-pair interaction capability. Our method not only guarantees the lowest training complexity under the premise of all-pair interaction capability but also does not require any pre-processing process, which shows superiority against the existing methods. All in all, our method achieves the strongest interaction capability with the most lightweight structure and the lowest complexity.

\section{Experiments}\label{sec5}

In the experimental section, we evaluate SpikeGraphormer on various datasets with different scales from different fields. Different scale datasets are helpful to analyze the model's computational efficiency and scalability to large-scale graphs. Different domain datasets can verify whether the model has general and effective cross-domain representation learning ability. We compare SpikeGraphormer with several advanced methods, including GNNs (GCN \cite{kipf2016semi}, GAT \cite{velivckovic2018graph}, SGC \cite{wu2019simplifying}) and advanced GNNs ($\text{H}_{2}\text{GCN}$ \cite{zhu2020beyond} and SIGN \cite{frasca2020sign}); Transformers-based methods (Graphormer \cite{ying2021transformers}, GraphTrans \cite{wu2021representing} ANS-GT \cite{zhang2022hierarchical} and Nodeformer \cite{wu2022nodeformer}). Among them, Nodeformer \cite{wu2022nodeformer} is the latest research in optimizing the computational complexity of Graph Transformers and is also the most competitive baseline method.

We first compare our method with baselines in terms of node classification performance and GPU memory cost on medium-scale (Section \ref{sec5.2}) and large-scale graphs (Section \ref{sec5.3}). In Section \ref{sec5.4}, to more clearly compare the computational efficiency and scalability of different methods, we first conduct a comparative analysis of the training time, inference time, and GPU memory cost on different datasets. In addition, we also visually show the changes in training time and GPU memory cost by constructing datasets containing different numbers of nodes. Aiming at the universality of the model, following Nodeformer \cite{wu2022nodeformer}, we explore the possibility of cross-domain application of the model on the image classification dataset Mini-ImageNet \cite{vinyals2016matching} and text classification dataset 20News-Groups \cite{pedregosa2011scikit}, respectively (Section \ref{sec5.5}). Finally, we carry out ablation experiments of the dual-branch architecture (Section \ref{sec5.6}) and hyperparameters experiments (Section \ref{sec5.7}), to verify the dependence on the graph structure information on different datasets, the model's robustness, and the setting suggestions of important parameters.

\begin{table*}[t] 
\center
  \caption{Node classification performance of different methods on medium-scale datasets. "OOM" indicates Out-Of-Memory during training with 24GB GPU memory. The best results are shown in bold and the second-best results are underlined.}
\label{tab2}
\begin{tabular}{cccccc}

\toprule

\textbf{Datasets} & \textbf{Chameleon}  & \textbf{Cora} & \textbf{Squirrel} & \textbf{Film} & \textbf{Deezer}    \\
\midrule
\textbf{\#nodes}        & 890        & 2708       & 2223        & 7600     & 28281     \\
\textbf{\#edges}        & 8845       & 5278       & 46998       & 29926    & 92752     \\
\midrule
GCN             & 41.4 $\pm$ 2.9  & 81.5 $\pm$ 0.5  & 38.8 $\pm$ 1.7  & 30.8 $\pm$ 0.4 & 63.2 $\pm$ 0.5  \\
GAT             & 39.6 $\pm$ 3.0  & \underline{82.4 $\pm$ 0.6}  & 36.2 $\pm$ 2.2  & 30.1 $\pm$ 0.6 & 62.3 $\pm$ 0.7  \\
SGC             & 39.2 $\pm$ 3.1 & 80.6 $\pm$ 0.3  & 39.6 $\pm$ 2.4  & 28.3 $\pm$ 1.1 & 62.9 $\pm$ 0.6  \\
$\text{H}_{2}\text{GCN}$          & 38.3 $\pm$ 3.9  & 82.0 $\pm$ 0.6  & 35.6 $\pm$ 1.3   & 34.6 $\pm$ 1.5 & 66.0 $\pm$ 0.7  \\
SIGN            & 41.9 $\pm$ 2.3 & 81.8 $\pm$ 0.4 & 40.6 $\pm$ 2.6  & 36.5 $\pm$ 1.1 & 66.2 $\pm$ 0.4  \\

\midrule
$ \text{Graphormer}_\text{small}$ & 41.8 $\pm$ 2.6  & 75.5 $\pm$ 1.2  & 40.7 $\pm$ 2.9   & OOM      & OOM       \\
$ \text{GraphTrans}_\text{small}$ & 42.1 $\pm$ 2.8  & 80.9 $\pm$ 0.9  & 40.9 $\pm$ 2.6   & 33.2 $\pm$ 0.8 & OOM       \\
ANS-GT          & \underline{42.6 $\pm$ 2.7}  & 82.3 $\pm$ 0.6   & \underline{41.1 $\pm$ 1.5}    & 35.9 $\pm$ 1.4 & \underline{66.6 $\pm$ 0.8} \\
Nodeformer      & 34.9 $\pm$ 4.1  & 82.1 $\pm$ 0.6 & 38.6 $\pm$ 1.0    & \underline{36.8 $\pm$ 1.2} & 66.3 $\pm$ 0.7 \\
\textbf{SpikeGraphormer} & \textbf{44.8 $\pm$ 4.2}  & \textbf{84.8 $\pm$ 0.7}   & \textbf{42.6 $\pm$ 2.4} & \textbf{38.1 $\pm$ 1.5} & \textbf{67.5 $\pm$ 0.8}
\\
\midrule
\end{tabular}

\end{table*}

\subsection{Basic Setup}\label{sec5.1}

Without special explanation, all experiments mentioned in this paper are conducted on a single NVIDIA RTX-4090 GPU with 24GB memory. For some models that cannot be trained in 24GB GPU memory, such as Graphormer and GraphTrans, we reduce the number of layers and attention headers to ensure trainability as much as possible. Certainly, there are some instances where even after these adjustments, the training still exceeds the available GPU memory, which we denote as "OOM" (Out-Of-Memory) in the results. Specifically, we utilize Binary Cross-Entropy (BCE) Loss for two-class classification tasks and Negative Log-Likelihood (NLL) Loss for multi-class classification problems. The parameters of the models are initially randomized, and the Adam optimizer is adopted for the gradient-based optimization of our model. We adopt the early-stop mechanism with patience=30, the maximum epoch is set to 1000. For all baselines, we refer to the original paper and conduct grid searches on the hyperparameters to determine their optimal parameter combination. For our method, the following are the search ranges for several important parameters: time steps of SNNs $T\in\{1,2,3,4\}$; $dropout\in\ \{0.1,0.2,0.3,0.4,0.5,0.6,0.7\}$; GNN branch weight $\alpha\in\ \{0.1,0.2,0.3,0.4,0.5,0.6,0.8\}$; SpikeGraphormer encoder blocks $L_1\in\{1,2,3,4\}$; graph structure encoder blocks $L_2\in\{1,2,3,4,5,6,7,8,9\}$; embedding dimension $\in\ \{32,64,128,256\}$; learning rate $lr\in\ \{1e-4,2e-4,5e-4,1e-3,2e-3,5e-3,1e-2,2e-2,5e-2,0.1\}$; attention\ headers $\in\{1,2,3,4\}$. To ensure the stability of the results, all the reported results are the average results of 10 experiments. We release our code and data at \url{https://github.com/PHD-lanyu/SpikeGraphormer}.

    \begin{table}[t]
\center
\caption{Performance and training GPU memory cost of different methods on OGB-Proteins dataset with batch size 10K. The best results are shown in bold and the second-best results are underlined.}
\label{tab3}
\begin{tabular}{ccc}
\toprule
\textbf{Method}          & \textbf{ROC-AUC(\%)}          & \textbf{Train Mem} \\
\midrule
MLP             & 72.04 $\pm$ 0.48         & 2.0 GB    \\
GCN             & 72.51 $\pm$ 0.35         & 2.5 GB    \\
SGC             & 70.31 $\pm$ 0.23         & 1.2 GB    \\
GraphSAINT-GCN  & 73.51 $\pm$ 1.31         & 2.3 GB    \\
GraphSAINT-GAT  & 74.63 $\pm$ 1.24         & 5.2 GB    \\
Nodeformer      & \underline{ 77.45 $\pm$ 1.15}   & 3.2 GB    \\
\textbf{SpikeGraphormer} & \textbf{79.62 $\pm$ 0.90} & 3.7  GB  \\
\midrule
\end{tabular}
\end{table}

\subsection{Results on Medium-sized Graphs}\label{sec5.2}

 In this section, we select 5 medium-sized datasets that are commonly used in graph representation learning. We refer to the previous work for the split of datasets. For the Cora dataset, we refer to GCN \cite{kipf2016semi}; for Film and Deezer, we refer to the original paper \cite{lim2021new}; for Squirrel and Chameleon, we refer to the latest paper \cite{platonov2023critical}\footnote{The results on some datasets are not consistent with Nodeformer \cite{wu2022nodeformer}, which is caused by different data splitting, please see our open-source code for details.}.
 
  \begin{table}[h]
\center
\caption{Performance and training GPU memory cost of different methods on Amazon2M dataset with batch size 100K. The best results are shown in bold and the second-best results are underlined.}
\label{tab4}
\begin{tabular}{ccc}
\toprule
\textbf{Method}                   & \textbf{Accuracy(\%) }        & \textbf{Train Mem }\\
\midrule
MLP                      & 63.46 $\pm$ 0.10         & 1.4 GB    \\
GCN                      & 83.90 $\pm$ 0.10         & 5.7 GB    \\
SGC                      & 81.21 $\pm$ 0.12         & 1.7 GB    \\
GraphSAINT-GCN           & 83.84 $\pm$ 0.42         & 2.1 GB    \\
GraphSAINT-GAT           & 85.17 $\pm$ 0.32         & 2.2 GB    \\
Nodeformer               & \underline{87.85 $\pm$ 0.24}          & 4.0 GB    \\
\textbf{SpikeGraphormer} & \textbf{88.12 $\pm$ 0.06} & 4.8 GB   \\
\midrule
\end{tabular}

\end{table}
 
 \begin{table*}[t]
\center
\caption{Efficiency comparison of Nodeformer and SpikeGraphormer w.r.t. training time, inference time, and GPU memory cost per epoch on RTX-4090 with 24GB memory.}
\label{tab5}
\setlength\tabcolsep{2.8pt}
  \renewcommand{\arraystretch}{1.0}
% \scriptsize
\begin{tabular}{c|ccc|ccc|ccc}
\toprule
\multirow{2}{*}{\textbf{Method}}  & \multicolumn{3}{c|}{\textbf{Cora}}   & \multicolumn{3}{c|}{\textbf{Squirrel}} & \multicolumn{3}{c}{\textbf{Chameleon}} \\
                         & \textbf{Tr(ms)} & \textbf{Inf(ms)} & \textbf{GPU(MB)} & \textbf{Tr(ms)}  & \textbf{Inf(ms)}  & \textbf{GPU(MB)} & \textbf{Tr(ms)}  & \textbf{Inf(ms)}  & \textbf{GPU(MB)}  \\
\midrule

Nodeformer               & 26.27  & 10.04   & 238.95  & 27.75   & 14.52    & 4159.99 & 45.36   & 13.62    & 126.36   \\
\textbf{SpikeGraphormer} & 23.34  & 9.32    & 92.8    & 21.49   & 5.55     & 132     & 31.55   & 9.98     & 75.87   \\
\midrule
\end{tabular}
 \vspace{-0.4cm}
\end{table*}
  Table \ref{tab2} shows the performance of different methods. We can intuitively see that our proposed SpikeGraphormer achieves the best performance on all datasets. SpikeGraphormer consistently outperforms the basic GNNs (GCN \cite{kipf2016semi}, GAT \cite{velivckovic2018graph}, and SGC \cite{wu2019simplifying}) by a large margin (up to 23.7\%). In comparison to advanced GNNs ($\text{H}_{2}\text{GCN}$ \cite{zhu2020beyond} and SIGN \cite{frasca2020sign}), SpikeGraphormer also shows up to 6.9\% performance improvement. In most cases, Transformer-based methods exhibit more competitive performance than GNNs. As we all know, GNN-based methods have some deficiencies in capturing long-range neighbors \cite{sun2023mhnf}. This reflects the benefits of the Transformer's ability to capture global interactions. What's more, SpikeGraphormer still maintains the superiority (up to 5.1\%) against the latest Transformers-based methods (Graphormer \cite{ying2021transformers}, GraphTrans  \cite{wu2021representing}, ANS-GT \cite{zhang2022hierarchical}, and Nodeformer \cite{wu2022nodeformer}). This challenges the prevailing perception that SNNs' performance is inherently inferior to traditional ANNs. Although we have simplified the model structure of Graphormer and GraphTrans as much as possible, their complex model design and quadratic complexity of attention layers also lead to the dilemma of "OOM" on Film and Deezer datasets. Thanks to the well-designed node sampling strategy and hierarchical attention scheme, ANS-GT not only reduces the GPU memory cost but also maintains a satisfactory performance. The power of all-pair node interactions is not demonstrated by Nodeformer, especially on Squirrel, Deezer, and Chameleon datasets. On the one hand, the scale of these graphs is still relatively small. On the other hand, Film, Squirrel, Deezer, and Chameleon are heterophilic graphs where neighboring nodes tend to have distinct labels, and the effect of integrating the graph structure information through additional edge loss may not be satisfactory.

\subsection{Results on Large-sized Graphs} \label{sec5.3}

In this section, we select two large-scale datasets, OGB-Proteins (0.13M nodes and 39M edges) and Amazon2M (2.4M nodes and 61M edges). Compared to OGB-Proteins, Amazon2M contains more nodes but has a more sparse structure. The two datasets are both from OGB \cite{hu2020open},  we employ a random splitting strategy, allocating 50\% of the nodes for training, 25\% for validation, and another 25\% for testing. Since OGB-proteins is a multi-task dataset with an output dimension of 112, following OGB \cite{hu2020open}, ROC-AUC is adopted as the evaluation metric. Unlike medium-size datasets, the datasets used in this section are too large to allow full-batch training on the entire graph with 24GB GPU memory. Consequently, for these larger datasets, we have to resort to alternative strategies such as mini-batch training to handle the GPU memory constraints. During the validation and testing phases, we revert to performing full graph inference and evaluations on the CPU. After the trade-off between performance and GPU memory, in the implementation, we set the batch size to 10K and 100K for OGB-Proteins and Amazon2M datasets, respectively.

% \usepackage[normalem]{ulem}

The performance and training GPU memory cost of different methods on OGB-Proteins and Amazon2M are shown in Table \ref{tab3} and Table \ref{tab4}, respectively. It can be seen intuitively from the tables that the SpikeGraphormer achieves the best performance while maintaining a relatively lower GPU memory cost. As can be seen from the tables, the GPU memory cost of our approach is even lower than that of GCN \cite{kipf2016semi}, because the GNN branch of SpikeGraphormer requires a relatively small number of GCN layers (1-2 layers), while the GCN requires more layers to achieve the desired performance. Unlike the previous situation on medium-size datasets, Nodeformer \cite{wu2022nodeformer} emerges as the most competitive baseline method when it comes to dealing with large-scale datasets. As we have previously analyzed, all-pair node interactions help Nodeformer and SpikeGraphormer capture more underlying relationships and global information, significantly boosting the performance of downstream tasks. To our surprise, SpikeGraphormer, an SNN-driven attention method, outperforms all traditional ANN-driven approaches, which greatly encourages the subsequent application of SNNs in graph representation learning.

\subsection{Efficiency and Scalability Comparison}\label{sec5.4}

To better quantify the efficiency and scalability of our proposed method on large-scale graphs, we record the training time, inference time, and GPU memory cost per epoch in this section. Here, we only select the most competitive Nodeformer for comparison, all the hyperparameters are the combinations (same as the settings in Section \ref{sec5.2}) when each method achieves the best performance within the search range. The experimental results are shown in Table \ref{tab5}. The table vividly illustrates that although the complexity of SpikeGraphormer and Nodeformer are both $O(N)$, SpikeGraphormer requires less training time and less inference time per epoch. Furthermore, its GPU memory consumption is far lower than that of Nodeformer. Combined with the performance results in Section \ref{sec5.2}, we can conclude that SpikeGraphormer achieves higher performance with less computing energy and GPU memory.

% \usepackage{multirow}

\begin{figure*}[htb]
  \centering
  \includegraphics[width=6in,trim=0 0 0 0]{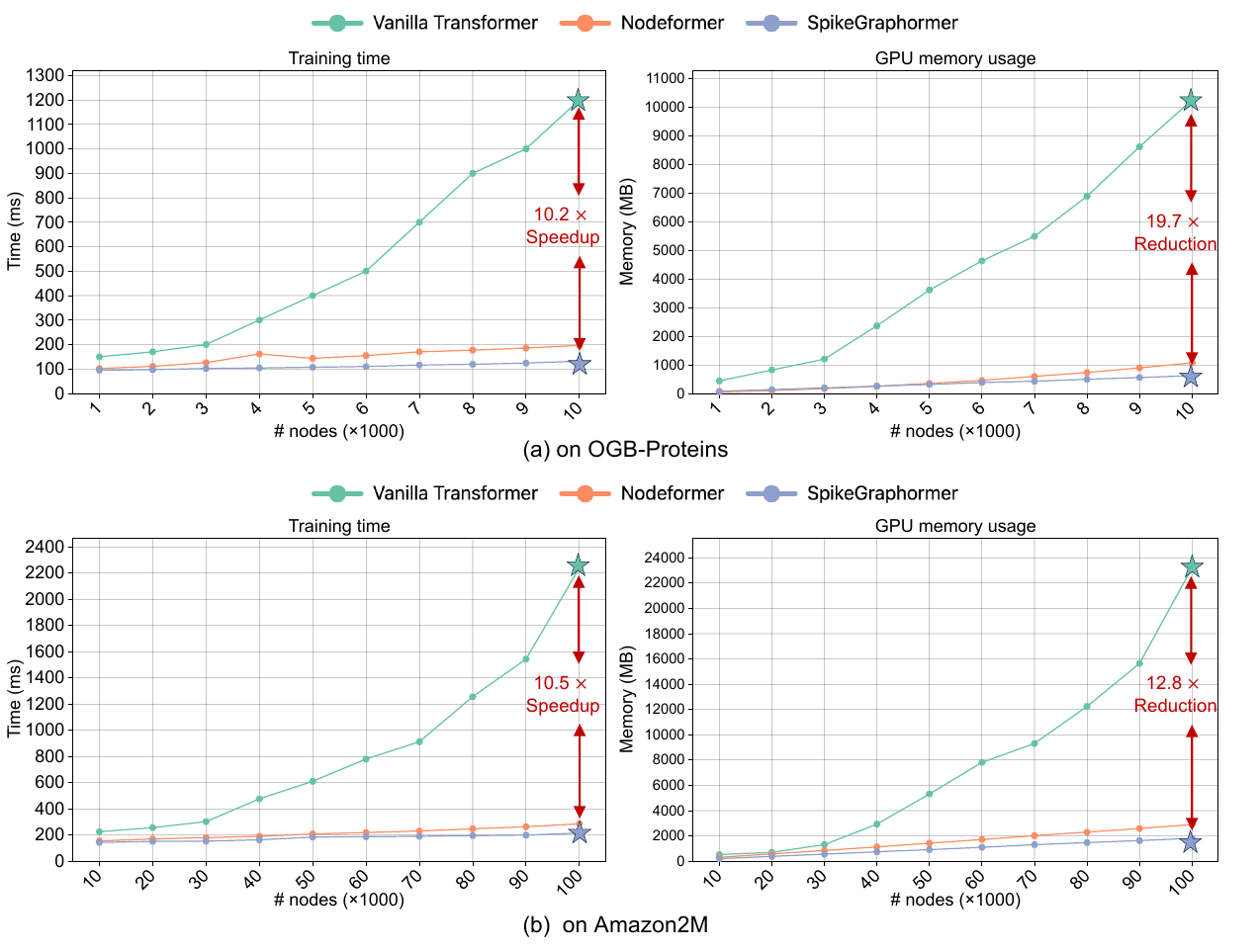}
  \caption{The trends in training time and GPU memory cost of the three methodologies under different number nodes.}
  \label{fig4}
%  \vspace{-0.2cm}
  \end{figure*}

To further explore the model's scalability, we randomly sample different numbers of nodes on OGB-Proteins and Amazon2M to build graphs of different scales for testing. To better reflect the gains of computing efficiency brought about by linear complexity, we build a Vanilla ANN-driven Transformer, which possesses the identical scale and structure as SpikeGraphormer. For a fair comparison, the hidden layer dimension of each method is uniformly set to 64. Fig.\ref{fig4} shows the changes in training time and GPU memory cost of the three methodologies under different number nodes. We can intuitively see the tremendous performance improvement brought by linear complexity. With the increase of nodes, the quadratic complexity makes the training time and GPU memory consumption of Vanilla Transformer rise sharply, while SpikeGraphormer and Nodeformer always maintain within the linear growth range due to their linear complexity (10 $\times$ $\sim$ 20 $\times$ lower than Vanilla Transformer). By comparing SpikeGraphormer and Nodeformer, we are surprised to find that SpikeGraphormer has better efficiency, and the advantage is more obvious when the graph scale expands, which indicates its better scalability on larger graphs.

%\begin{figure}[htb]
%%\setlength{\belowcaptionskip}{-1.cm}
%%\setlength{\abovecaptionskip}{-0.1cm}
%  \centering
%  %\includegraphics[width=3.5in,trim=0 40 0 20]{figures/figure2.pdf}
%  \includegraphics[width=3.3in,trim=0 0 0 0]{over-smoothing-metric-hebing.pdf}
%  \caption{Quantitative performance analysis of different methods with different \#layers/hops settings on different datasets.}
%  \label{fig5}
%  \vspace{-0.2cm}
%  \end{figure}

\subsection{Generalization to Image and Text Classification}\label{sec5.5}

In this section, we verify the generality of SpikeGraphormer in the field of image and text classification. Referring to Nodeformer \cite{wu2022nodeformer}, we select the image dataset Mini-ImageNet \cite{vinyals2016matching} and the text classification dataset 20News-Groups \cite{pedregosa2011scikit}. 20News-Groups contains 20,000 newsgroup documents, these documents are split into 20 classes. In our setup, we select 10 classes of documents from the dataset and utilize words (represented by their TF-IDF scores) with a frequency threshold of over 5\% as features for classification. Mini-ImageNet contains 100 categories of images, each category includes 600 samples, each of which is an 84 $\times$ 84 RGB image. In the experiment, we utilize the processed datasets by Nodeformer \cite{wu2022nodeformer}, which selects 30 types of images. The 128-dimensional image features are obtained by CNN. The above two original datasets do not contain any network structure, that is, there is no edge between nodes. Here, we input node features into $k$-NN to construct the graph. In the experiment, we determine $k$ as 5,10 and 15.

Table \ref{tab6} shows the performance of different methods under different values of $k$. To our surprise, our method still achieves the most competitive performance on image and text classification tasks, which indicates its fantastic cross-domain generalization. By comparing the results under different $k$ values, we observe that only SpikeGraphormer and Nodeformer consistently maintain stability in performance, while other methods show fluctuations, especially on the 20News-Groups dataset. There is no discernible decline or rise in performance for SpikeGraphormer and Nodeformer even when the graph structure information is not incorporated. This observation suggests that on these two datasets, SpikeGraphormer and Nodeformer do not rely heavily on explicit graph structure information to achieve strong performance. Instead, their inherent capability for all-pair node interactions has allowed them to effectively capture potentially valuable relationships among the nodes.

% \usepackage{multirow}
\begin{table*}[t]
\center
\caption{Experimental results of image classification and node classification tasks on Mini-ImageNet and 20News-Groups respectively. $k$-NNs with different $k$ values are leveraged to construct the graph, and "w/o graph" means that no graph is used.}
\label{tab6}
 \scriptsize
\begin{tabular}{c|cccc|cccc}
\toprule
\multirow{2}{*}{\textbf{Method}}     & \multicolumn{4}{c|}{\textbf{Mini-ImageNet}}                                                    & \multicolumn{4}{c}{\textbf{20News-Groups}}                                                             \\
                                     & k=5                   & k=10                  & k=15                  & k=20                  & k=5                   & k=10                  & k=15                  & k=20                  \\
                                     
\midrule

GCN                                  & 84.86 $\pm$ 0.42          & 85.61 $\pm$   0.40        & 85.93 $\pm$   0.59        & 85.96 $\pm$   0.66        & 65.98 $\pm$   0.68        & 64.13 $\pm$   0.88        & 62.95 $\pm$   0.70        & 62.59 $\pm$   0.62        \\
GAT                                  & 84.70 $\pm$   0.48        & 85.24 $\pm$   0.42        & 85.41 $\pm$   0.43        & 85.37 $\pm$   0.51        & 64.06 $\pm$   0.44        & 62.51 $\pm$   0.71        & 61.38 $\pm$   0.88        & 60.80 $\pm$   0.59        \\
DropEdge                             & 83.91 $\pm$   0.24        & 85.35 $\pm$   0.44        & 85.25 $\pm$ 0.63          & 85.81 $\pm$   0.65        & 64.46 $\pm$   0.43        & 64.01 $\pm$   0.42        & 62.46 $\pm$   0.51        & 62.68 $\pm$   0.71        \\
IDGL                                 & 83.63 $\pm$   0.32        & 84.41 $\pm$   0.35        & 85.50 $\pm$   0.24        & 85.66 $\pm$   0.42        & 65.09 $\pm$   1.23        & 63.41 $\pm$   1.26        & 61.57 $\pm$   0.52        & 62.21 $\pm$   0.79        \\
LDS                                  & OOM                   & OOM                   & OOM                   & OOM                   & \textbf{66.15 $\pm$ 0.36} & 64.70 $\pm$   1.07        & 61.57 $\pm$   0.52        & 62.21 $\pm$   0.79        \\
Nodeformer                           & 86.77 $\pm$   0.45        & 86.74 $\pm$   0.23        & 86.87 $\pm$   0.41        & 86.64 $\pm$   0.42        & 66.01 $\pm$   1.18        & 65.21 $\pm$   1.14        & 64.69 $\pm$   1.31        & 64.55 $\pm$   0.97        \\
\textbf{SpikeGraphormer}             & \textbf{86.88 $\pm$ 0.42} & \textbf{86.82 $\pm$ 0.42} & \textbf{86.90 $\pm$ 0.39} & \textbf{86.68 $\pm$ 0.42} & 65.75 $\pm$   0.62        & \textbf{65.32 $\pm$ 0.76} & \textbf{64.76 $\pm$ 1.04} & \textbf{64.89 $\pm$ 0.97} \\
\midrule

Nodeformer w/o graph                 & \multicolumn{4}{c|}{\colorbox{lightgray}{\textbf{87.46 $\pm$   0.36}}}                                                   & \multicolumn{4}{c}{64.71 $\pm$   1.33}                                                            \\
\textbf{SpikeGraphormer   w/o graph} & \multicolumn{4}{c|}{86.84 $\pm$ 0.84}                                                              & \multicolumn{4}{c}{\colorbox{lightgray}{\textbf{65.50 $\pm$   1.05}}}             \\

\midrule                                     
\end{tabular}
\end{table*}

\begin{figure*}[h]
  \centering
  \includegraphics[width=7in,trim=0 0 0 0]{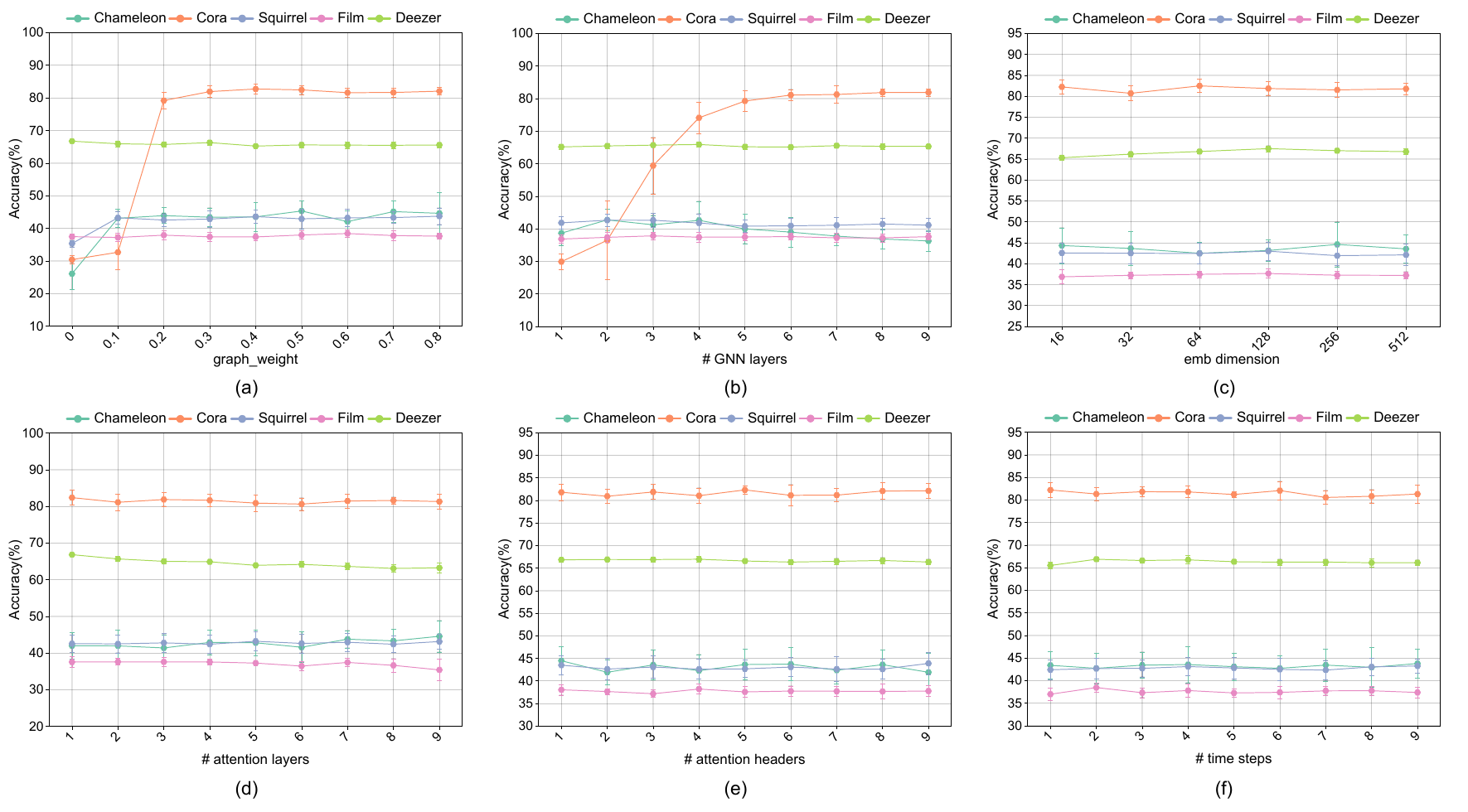}
  \caption{Performance of SpikeGraphormer with different parameter settings on different datasets.}
  \label{fig5}
  \vspace{-0.2cm}
  \end{figure*}

\subsection{Ablation Study of Dual-branch Architecture}\label{sec5.6}

In the recent advancements in Graph Transformer research, incorporating graph structure information has emerged as an essential component. In this paper, we design a Dual-branch architecture and leverage a simple yet surprisingly effective fusion strategy (addition or concatenation), to integrate the output of sparse GNN and SGA-driven Graph Transformer branches. Therefore, in this part, based on the optimal parameter combination in Section \ref{sec5.2}, we reset different graph weights ($\alpha$) and different GNN layers ($L_2$) to investigate the effectiveness of the Dual-branch architecture and the contribution of graph structures on different datasets.

The experimental results are shown in Fig.\ref{fig5} (a) and Fig.\ref{fig5} (b) respectively. One of the most obvious phenomena is that when $\alpha$ is set to 0, indicating a complete absence of graph structure information, we observe unsatisfactory performance across all datasets. This finding underscores that even though some datasets might be less sensitive to the different values of graph weight, the Dual-branch Architecture is effective and indispensable, and the incorporation of graph structure information remains essential for achieving optimal results.

On the Cora dataset, the model's performance notably deteriorates when $\alpha$ is set to a value lower than 0.2, and the same situation will occur when $L_2$ is less than 5. On other datasets, however, the model's performance does not show sensitivity to variations of these two parameters. We believe that the inherent nature of the datasets contributes to this phenomenon. Cora is a highly homogeneous graph where neighbor nodes have a high probability of sharing the same label. Other datasets are heterophilic graphs, where the neighbor nodes tend to possess different labels. Therefore, on the Cora dataset, the model needs more local graph structure information to help it obtain better label prediction results. For the datasets that do not contain graph structures, such as Mini-ImageNet and 20News-Groups in Section \ref{sec5.5}, SpikeGraphormer consistently maintains stability whether there is graph structure information or not, which again verifies our conclusion. Therefore, in practical application scenarios, properly setting different graph weight ($\alpha$) and different GNN layers ($L_2$) according to specific data characteristics is an important step to well-release the model's performance.

\subsection{Hyper-parameter Studies}\label{sec5.7}

Finally, we explore the influence of several important hyperparameters on SpikeGraphormer's performance, including embedding dimension, attention layers, attention headers, and time steps ($T$) of SNNs. The experimental results are shown in Fig.\ref{fig5} (c) - (f). To our surprise, SpikeGraphormer exhibits outstanding stability across various hyperparameter configurations. This remarkable stability lays a solid foundation for the robust running of SpikeGraphormer in real-world application scenarios and reduces the potential challenges faced by practitioners during deployment. Most importantly, we can set smaller time steps ($T$), embedding dimension, attention layers, and attention headers while maintaining excellent performance or with negligible decline, further reducing computation energy and memory cost. In this way, the scalability of the model for large-scale graphs can be promoted again.

\section{Conclusions}\label{sec6}

This paper addresses the long-standing challenge of computational complexity and GPU memory explosion that has troubled Graph Transformers, by introducing bio-inspired SNNs to optimize the efficiency of self-attention computation. First, we propose the Spiking Graph Attention (SGA) module, benefiting from the unique event-driven and binary spike characteristics of SNNs, SGA can reduce the complexity from quadratic to linear of the number of nodes. This effort makes it possible to realize all-pair node interactions on large-scale graphs with limited computing power and GPU memory. Second, to relieve the inherent deficiency of Transformers in capturing local graph structures, we design a novel Dual-branch architecture named SpikeGraphormer, the simple yet surprisingly effective fusion strategy enables SpikeGraphormer to simultaneously perform all-pair node interactions and capture local neighborhoods. We have carried out experimental validations on several graph datasets with different scales and characteristics, and the experimental results show that our method achieves the best or most competitive performance when compared to state-of-the-art models. What's more, our method shows great potential in reducing training time, inference time, and GPU memory consumption, which is 10 $\times$ $\sim$ 20 $\times$ lower than vanilla self-attention. We apply our model to image and text classification, and the excellent performance also demonstrates its cross-domain generality.

Of course, our method is the first attempt to introduce SNNs into Graph Transformers, and there is still room for improvement. As we all know, SNNs possess a natural advantage in mining temporal features. In the future, we will continue to explore the application of SNNs in dynamic graphs where nodes, edges, or their features change over time \cite{li2023scaling}. At the same time, the remarkable performance of our model in the field of image and text classification also encourages us to continue to explore more application areas.

% or


%\appendices

%\section{}

\ifCLASSOPTIONcompsoc
  \section*{Acknowledgments}
\else
  \section*{Acknowledgment}
\fi

This work is supported by the National Key R\&D Program of China (2020YFE0201500), the Fundamental Research Funds for the Central Universities (HIT.NSRIF.201714), Weihai Science and Technology Development Program (2016DXGJMS15), and the Key Research and Development Program in Shandong Province (2017GGX90103).
This work is supported by the National Key R\&D Program of China (2020YFE0201500), Weihai Scientific Research and Innovation Fund (2020), and Shandong Province science and technology smes innovation ability improvement project (2023TSGC0695).

 Our code is built upon Nodeformer \cite{wu2022nodeformer} and  Spike-driven Transformer \cite{yao2024spike}, we thank the authors for their open-sourced code.

\ifCLASSOPTIONcaptionsoff
  \newpage
\fi

\bibliography{cas-refs}

%\begin{IEEEbiography}[{\includegraphics[width=1in,height=1.25in,clip,keepaspectratio]{Yundong_Sun.jpg}}]{Yundong Sun}
%  is currently working toward the Ph.D degree in School of Astronautics at Harbin Institute of Technology. His research interests include machine learning, graph representation learning, recommender system and network embedding.
%\end{IEEEbiography}
%\begin{IEEEbiography}[{\includegraphics[width=1in,height=1.25in,clip,keepaspectratio]{Dongjie_Zhu.jpg}}]{Dongjie Zhu}
%  received the Ph.D degree in computer architecture from the Harbin Institute of Technology, in 2015. He is an associate professor in School of Computer Science and Technology at Harbin Institute of Technology, Wehai. His research interests include parallel storage systems, social computing, machine learning.
%\end{IEEEbiography}
%
%
%\begin{IEEEbiography}[{\includegraphics[width=1in,height=1.25in,clip,keepaspectratio]{Haiwen_Du.jpg}}]{Haiwen Du}
%  is currently working toward the Ph.D degree in School of Astronautics at Harbin Institute of Technology. His research interests include storage system architecture and massive data management.
%\end{IEEEbiography}
%
%\begin{IEEEbiography}[{\includegraphics[width=1in,height=1.25in,clip,keepaspectratio]{Zhaoshuo_Tian.jpg}}]{Zhaoshuo Tian}
%  is a professor in School of Astronautics at Harbin Institute of Technology. His research interests include laser technology and marine laser detection technology. He is a member of IEEE.
%
%\end{IEEEbiography}
\end{document}